\documentclass[10pt,twocolumn,letterpaper]{article}

\usepackage{cvpr}              %

\usepackage{multirow}
\usepackage{xcolor,colortbl}
\definecolor{lavendergray}{rgb}{0.77, 0.76, 0.82}
\definecolor{lightgray}{rgb}{0.83, 0.83, 0.83}
\newcommand{\default}[1]{{\cellcolor{lightgray}#1}}
\newcommand{\compare}[1]{{\cellcolor{Gray}#1}}

\usepackage{array}
\newcolumntype{H}{>{\setbox0=\hbox\bgroup}c<{\egroup}@{}}
\newcommand{\tablestyle}[2]{\setlength{\tabcolsep}{#1}\renewcommand{\arraystretch}{#2}\centering\small}

\newcommand{\myparagraph}[1]{\vspace{0pt}\noindent{\bf #1}}

\usepackage{amssymb}%
\usepackage{pifont}%
\newcommand{\cmark}{\ding{51}}%
\newcommand{\xmark}{\ding{55}}%

\newcommand{\higherbetter}{$_\uparrow$}
\newcommand{\lowerbetter}{$_\downarrow$}

\newcommand{\ours}{{QLIP}\xspace}
\newcommand{\ourlm}{UM$^3$\xspace}

\def\cD{{\mathcal{D}}}
\def\cE{{\mathcal{E}}}
\def\cG{{\mathcal{G}}}
\def\cL{{\mathcal{L}}}
\def\cQ{{\mathcal{Q}}}
\def\cF{{\mathcal{F}}}

\def\sg{{\mathrm{sg}}}
\def\sign{{\mathrm{sign}}}

\def\sB{{\mathcal{B}}}

\DeclareMathOperator*{\argmin}{arg\,min}

\def\vc{{\bm{c}}}
\def\ve{{\bm{e}}}
\def\vu{{\bm{u}}}
\def\vv{{\bm{v}}}
\def\vw{{\bm{w}}}
\def\vx{{\bm{x}}}
\def\vz{{\bm{z}}}

\usepackage{amssymb,bm}
\def\mC{{\bm{C}}}
\def\mH{{\bm{H}}}
\def\mX{{\bm{X}}}
\def\mY{{\bm{Y}}}
\def\mZ{{\bm{Z}}}

\newcommand{\E}{\mathbb{E}}
\newcommand{\R}{\mathbb{R}}

\def\Eqsref#1{Eq~\eqref{#1}}

\usepackage{wrapfig}

\definecolor{myblue}{rgb}{0.11764705882352941, 0.5647058823529412, 1.0}
\definecolor{Gray}{gray}{0.9}
\definecolor{darkgreen}{rgb}{0.545, 0.749, 0.608}
\newcommand{\gain}[1]{{\color{myblue}\scriptsize\textbf{#1}}}
\newcommand{\loss}[1]{{\color{darkgreen}\scriptsize\textbf{#1}}}
\newcommand{\basenum}[1]{{\color{gray}\scriptsize\textbf{#1}}}

\usepackage{tablefootnote}

\usepackage{soul}
\sethlcolor{Gray}

\definecolor{cvprblue}{rgb}{0.21,0.49,0.74}
\usepackage[pagebackref,breaklinks,colorlinks,allcolors=cvprblue]{hyperref}

\def\confName{CVPR}
\def\confYear{2025}

\title{
\ours: Text-Aligned Visual Tokenization Unifies Auto-Regressive Multimodal Understanding and Generation
}

\author{Yue Zhao$^{1,}$\thanks{Work done during an internship at NVIDIA Research.}
\hspace{0.2cm}
Fuzhao Xue$^{2,}$\thanks{Now at Google DeepMind.}
\hspace{0.2cm}
Scott Reed$^{2}$
\hspace{0.2cm}
Linxi Fan$^{2}$
\hspace{0.2cm}
Yuke Zhu$^{1,2}$
\hspace{0.2cm}
\\
Jan Kautz$^{2}$
\hspace{0.2cm}
Zhiding Yu$^{2}$
\hspace{0.2cm}
Philipp Kr\"ahenb\"uhl$^{1}$
\hspace{0.2cm}
De-An Huang$^{2}$
\\[.5ex]
$^1$UT Austin \hspace{0.5cm}
$^2$NVIDIA \hspace{0.5cm}
\\
\small{\url{https://nvlabs.github.io/QLIP/}}
}

\begin{document}
\maketitle

\begin{abstract}
We introduce Quantized Language-Image Pretraining (\textbf{\ours}), a visual tokenization method that combines state-of-the-art reconstruction quality with state-of-the-art zero-shot image understanding. 
\ours trains a binary-spherical-quantization-based autoencoder with reconstruction and language-image alignment objectives.
We are the first to show that the two objectives do not need to be at odds.
We balance the two loss terms dynamically during training and show that a two-stage training pipeline effectively mixes the large-batch requirements of image-language pre-training with the memory bottleneck imposed by the reconstruction objective.
We validate the effectiveness of \ours for multimodal understanding and text-conditioned image generation with a single model.
Specifically, \ours serves as a drop-in replacement for the visual encoder for LLaVA and the image tokenizer for LlamaGen with comparable or even better performance.
Finally, we demonstrate that \ours enables a unified mixed-modality auto-regressive model for understanding and generation.

\end{abstract}

\section{Introduction}
\label{sec:intro}

Auto-regressive sequence modeling and its variants have become the state-of-the-art paradigm for natural language modeling~\cite{achiam2023gpt4,dubey2024llama3}, multi-modal understanding~\cite{team2023gemini,liu2024llava1.5}, and arguably visual generation~\cite{yu2024magvit2,tian2024var}.
Despite encouraging progress, a unified auto-regressive model that performs well from any to any modality~\cite{wu2024nextgpt,lu2024unifiedio2,team2024chameleon} has proven difficult to train. 
One key issue lies in visual tokenization.
Commonly, an auto-encoder learns to reconstruct the input image with a set of visual tokens and leaves the joint visual-language modeling to the auto-regressive model.
This leads to tokenization that compresses the inputs visually, but not semantically, and consecutively leads to the two modalities competing and slow training~\cite{team2024chameleon}.

\begin{figure}
    \centering
    \includegraphics[width=0.9\linewidth]{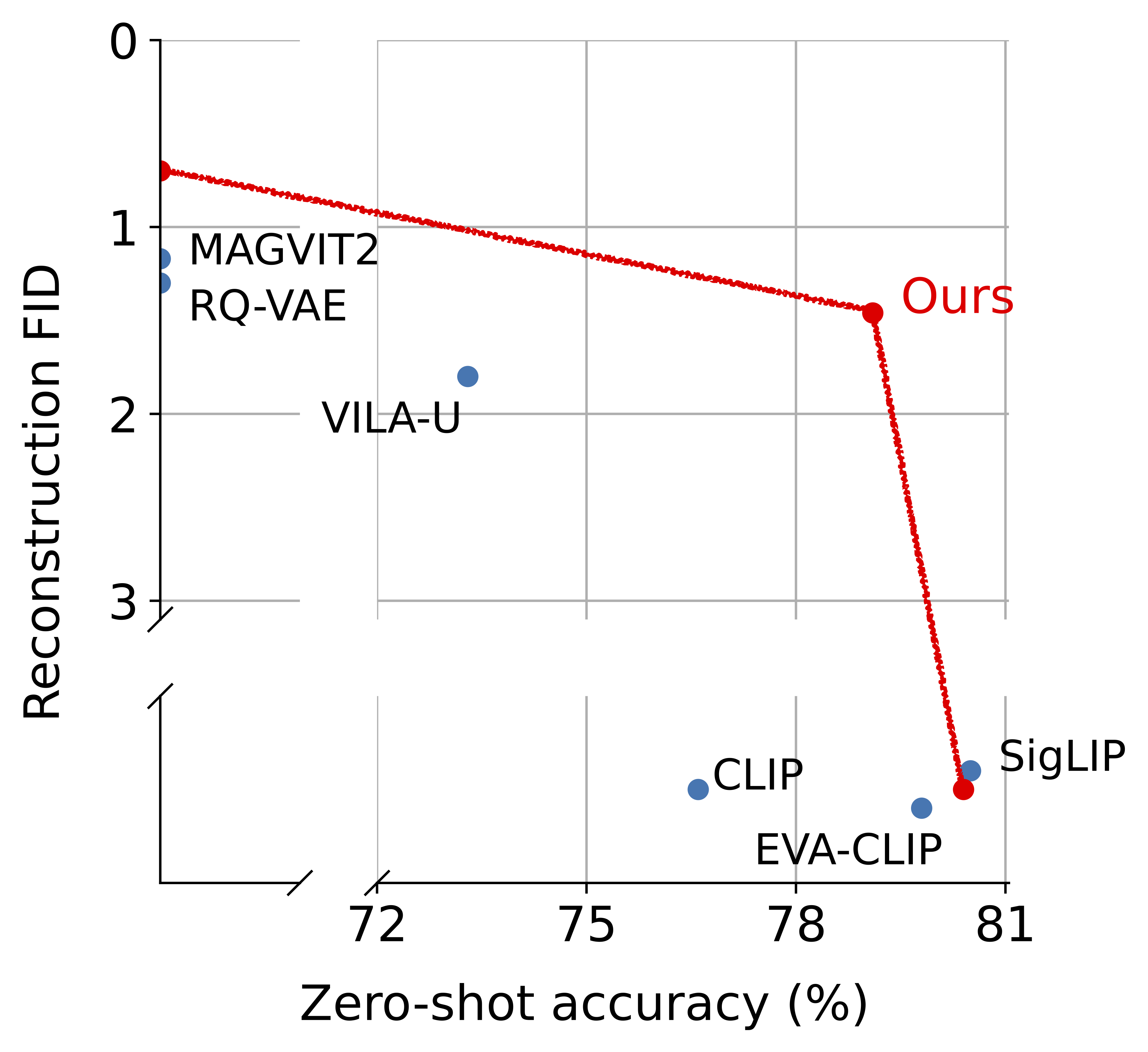}
    \vspace{-5pt}
    \caption{State-of-the-art visual tokenizers excel at either understanding (high zero-shot accuracy,~\eg SigLIP~\cite{zhai2023siglip}) or reconstruction (low reconstruction FID,~\eg MAGVIT2~\cite{yu2024magvit2}), but not both.
    \ours can perform well on both understanding and reconstruction with a marginal performance drop, opening up an opportunity for unified multi-modal understanding and generation.
    }
    \label{fig:teaser}
\end{figure}

In this paper, we propose to perform multi-modal alignment as early as the visual tokenization phase.
The result is a generic visual tokenizer for multi-modal language modeling that excels at capturing semantics and reconstructs high-quality visuals at the same time. 
We train a Binary Spherical Quantization (BSQ)-based Auto-encoder with a text-aligned visual-encoder through a contrastive objective.
We term the framework Quantized Language-Image Pretraining, \ours for short.

We identify two \textbf{main challenges} when training \ours.
First, contrastive alignment and regression objectives compete and are hard to balance.
Second, contrastive learning relies on large-batch training, while reconstruction losses incur a heavy memory cost~\cite{zhang2018lpips,karras2020stylegan2} and thus allow for only small batches.
To handle the first challenge, we observe the stark difference in the gradient magnitude leads to different convergence rates between the contrastive image-text alignment and pixel reconstruction objectives.
We introduce a simple and effective automated weighting scheme between the two losses.
We weigh the loss terms by the inverse of their post-hoc loss values without needing any extra cost to compute the gradient.
To handle the second challenge, we propose a two-stage training recipe.
In the first stage, we train \ours with a combination of alignment loss and MSE loss with memory-efficient Transformer architecture~\cite{dao2022flashattention,chen2016checkpointing,rajbhandari2020zero}.
In the second stage, we drop the text encoder, freeze the visual encoder, and no longer optimize the contrastive loss.
This allows for a smaller batch size and enables fine-tuning of just the bottleneck quantizer and the decoder using a weighted sum of MSE, perceptual loss, and generative adversarial (GAN) loss.

We empirically show that \ours achieves competitive reconstruction results compared to cutting-edge visual tokenizers, including continuous tokenizer (SD-VAE) and discrete tokenizer (BSQViT) under a similar compression ratio.
At the same time, \ours yields visual-text alignment capability similar to a CLIP-only objective.
Furthermore, we validate the effectiveness of our \ours tokenizer on a wide spectrum of multimodal understanding and generation benchmarks.
On LLaVA-based multimodal models, \ours shows a marginal loss of performance compared to the CLIP-only baseline under a fair comparison (\eg same input resolution and same instruction-tuning data). 
This is in contrast to the prior belief that vision tokenizers lead to substantial degradation when used in VLMs.
On text-conditioned image generation, \ours shows improved generation FID and better text-image alignment qualitatively compared to the language-agnostic visual tokenizer (VQ-VAE and BSQViT).
Finally, \ours enables a unified mixed-modal auto-regressive model that can handle language-only, image-to-text, and text-to-image tasks in a single model.

\section{Related Work}
\label{sec:related}

\myparagraph{Visual Tokenzation}.
Analogous to LLM tokenizers~\cite{sennrich2016bpe,schuster2012wordpiece,kudo2018sentencepiece} that losslessly transform a text string into discrete tokens, visual tokenization aims to map an image or video to tokens while keeping as much visual information as possible.
VQ-VAE~\cite{van2017vqvae} introduced the concept of discrete tokenized bottlenecks in auto-encoder architectures.
Later improvements include better training objectives~\cite{esser2021vqgan,ramesh2021dalle}, increasing VQ codebook usage~\cite{yu2022vitvqgan,zheng2023cvq}, and advanced quantization techniques~\cite{lee2022rqvae,mentzer2023fsq,yu2024magvit2,zhao2024bsq}.
All of these efforts aim for improved reconstruction quality using the same compression budget and benefit visual generation~\cite{chang2022maskgit,yu2024magvit2,tian2024var}.
However, better reconstruction quality does not necessarily lead
to better visual representation~\cite{he2022mae,wei2023diffmae}.
On the other hand, visual tokens serve as good intermediate supervision to learn visual encoders with strong representation~\cite{bao2022beit,peng2022beitv2,zhou2022ibot,li2023mage}.
Our work shows that by properly adding textual supervision the visual tokenizer can be a strong visual encoder \emph{without} introducing extra parameters. 
The concept of aligning visual tokenizer with language is also related to LQAE~\cite{liu2023lqae} and SPAE~\cite{yu2023spae}.
SPAE~\cite{yu2023spae} aligns the raw pixels with the language token embeddings from a frozen LLM directly.
However, SPAE needs more tokens to reconstruct comparably well with VQ-VAE, indicating that the frozen language codebook might not be optimal.

\myparagraph{Unifying understanding and generation.}
Visual tokenization enables unifying multi-modality in the same token space~\cite{lu2022unifiedio,lu2024unifiedio2,wu2024nextgpt,jin2024lavit,wang2024emu3,team2024chameleon,zhou2024transfusion}.
Chameleon~\cite{team2024chameleon} interleaves discrete visual and text tokens with a single Transformer and reported training difficulties.
Transfusion~\cite{zhou2024transfusion} combines text token prediction with diffusion for images.
Show-o~\cite{xie2024showo} unifies understanding and generation by masked language modeling but uses different tokenizers for different tasks.
We use an auto-regressive objective to handle both modalities and \ours enables quick visual-language adaptation from a pre-trained LLM.
Another line of works is \emph{encoder-free}~\cite{fuyu-8b,diao2024eve}, which maps patches of raw pixels into embeddings for joint visual-language modeling.
However, this approach is much less data-efficient~\cite{beyer2024paligemma} and unable to generate visual content.
VILA-U~\cite{wu2024vilau} is closely relevant in that its tokenizer is initialized from SigLIP~\cite{zhai2023siglip}.
However, the understanding performance drops drastically after re-training (see Figure~\ref{fig:teaser}).
Finally, our visual tokenizer takes advantage of textual supervision and pixel-level reconstruction, echoing recent studies that a mixture of expert vision encoders complement each other for vision-language understanding~\cite{tong2024cambrian,shi2024eagle}.

\begin{figure*}[!tb]
    \centering
    \includegraphics[width=1.0\linewidth]{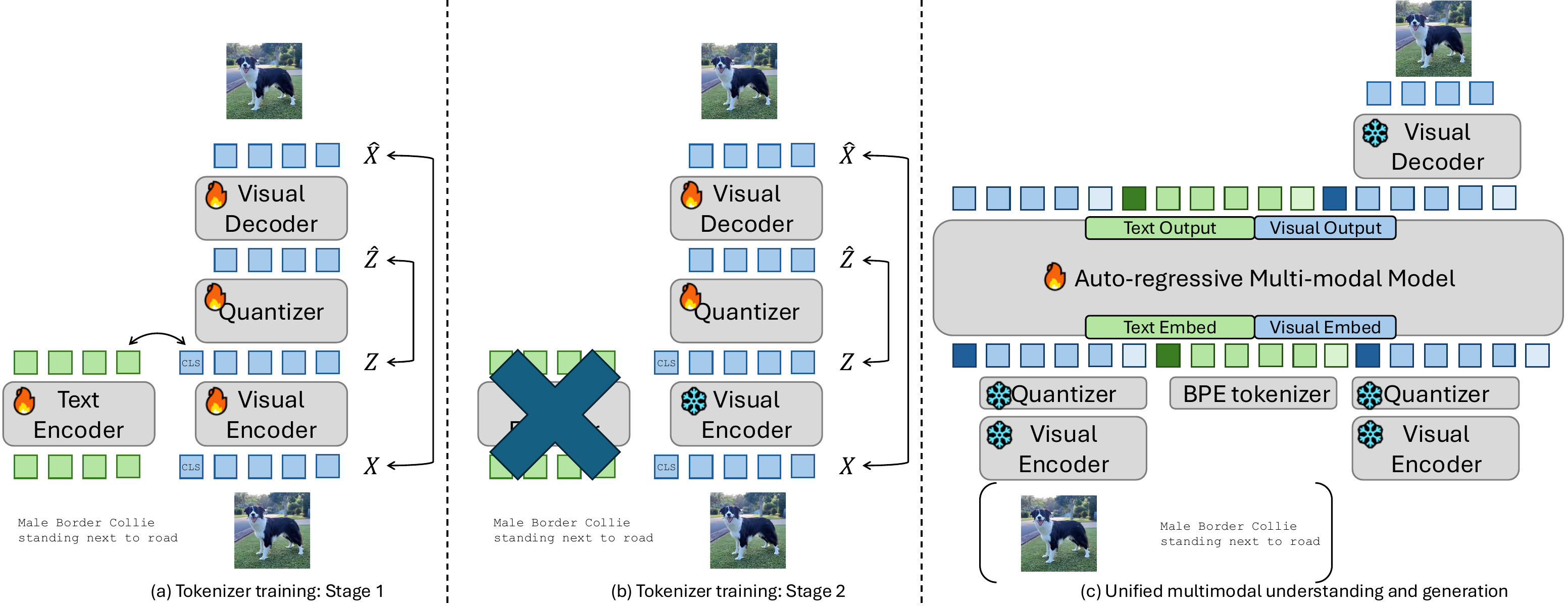}
    \vspace{-10pt}
    \caption{
    \textbf{Overview.}
    \textbf{(a-b)} Two-stage training pipeline of \ours.
    \textbf{(a)} In Stage 1, we train \ours with a combination of alignment loss and MSE loss.
    \textbf{(b)} In Stage 2, we drop the text encoder, freeze the visual encoder, and no longer optimize the contrastive loss.
    Only the bottleneck quantizer and the decoder are fine-tuned.
    \textbf{(c)} With the text-aligned visual tokenizer, we transform the image into visual tokens, concatenate them with text tokens, and use an auto-regressive multi-modal model (Sec~\ref{sec:method:um3}) to model jointly.
    }
    \label{fig:overview}
\end{figure*}

\section{Preliminaries}
\label{sec:prelim}

\myparagraph{Visual Tokenization} transforms an image to a set of \emph{discrete} tokens, which are later used for compression, generation, multi-modal understanding~\cite{zhao2024bsq,chang2022maskgit,team2024chameleon} via auto-regressive sequence modeling.
It has three basic components: a visual encoder $\cE$, a quantization bottleneck $\cQ$, and a visual decoder $\cG$.
Given an input image $\mX\in\R^{H\times W\times 3}$, the visual encoder $\cE$ produces a grid of $d$-dimensional latent embeddings $\mZ =\cE(\mX)\in\R^{\left(\frac{H}{p}\times\frac{W}{p}\right)\times d}$ downsampled by a factor $p$.
The bottleneck $\cQ$ transforms the real-valued latent embeddings into discrete tokens $\{\vc_1\ldots \vc_K\}$ in an element-wise fashion: ${\hat{\mZ} = \cQ(\mZ)\in\{\vc_1\ldots \vc_K\}^{\left(\frac{H}{p}\times\frac{W}{p}\right)}}$.
Finally, the decoder $\cG$ maps the discretized tokens back to the raw pixel space $\hat{\mX} = \cG(\hat{\mZ})\in\R^{H\times W\times 3}$.
The entire network $(\cE, \cG, \text{and }\cQ)$ is end-to-end trainable by minimizing a weighted sum of MSE loss $\cL_\mathrm{mse}=\|\hat{\mX}-\mX\|_2$, quantization loss $\cL_q(\cQ)$, and regularization terms,~\eg a commitment loss~\cite{van2017vqvae}, or perceptual and adversarial losses~\cite{esser2021vqgan}.

Vector Quantization (VQ)~\cite{van2017vqvae} $\cQ_\mathrm{VQ}$ maps latent inputs $\vz\in\mZ$ to the closest entry in a learnable codebook ${\mC=[\vc_1,\cdots, \vc_K]\in\R^{K\times d}}$:  ${\cQ_\mathrm{VQ}(\vz)=\argmin_{\vc_k\in\mC} \| \vz - \vc_k \|_2}$.
It uses the straight-through estimator (STE)~\cite{bengio2013ste} to propagate gradients through the quantization bottleneck.
Empirically, VQ scales poorly with increasing vocabulary size $K$~\cite{yu2024magvit2}.

Binary Spherical Quantization (BSQ)~\cite{zhao2024bsq} and Look-up Free Quantization (LFQ)~\cite{yu2024magvit2} provide a more scalable alternative.
They optimize an \emph{implicit} codebook. For example BSQ projects a hypercube onto a unit sphere and uses the corners of the hypercube as code vectors $\mC_\mathrm{BSQ}=\{-\frac{1}{\sqrt{L}}, \frac{1}{\sqrt{L}}\}^L$.
Each corner $\vc_k \in \mC_\mathrm{BSQ}$ corresponds to a
unique token $k$.
BSQ linear-projects the $d$-dimensional latent embedding $\vz$ to a $L$-dimensional unit hypersphere $\vu\in S^{L-1}$, applies binary quantization per axis $\hat{\vu} = \frac{1}{\sqrt{L}}\sign(\vu)$,
and back-projects to a quantized vector in the original latent space $\hat{\vz}$.
The code index at inference is obtained through binarization $ k=\sum_{i=1}^L 1_{[u_i>0]}2^{i-1} $.

To optimize for an effective latent code and encourage usage of the implicit codebook, the quantization loss uses an entropy objective~\cite{yu2024magvit2,jansen2020coincidence}
\begin{align}
\cL_\mathrm{BSQ} = \E\left[H(\cQ(\vz))\right] - \gamma H(\E[\cQ(\vz)]),
\label{eq:bsq}
\end{align}
where both entropy terms rely on a soft quantization~\cite{agustsson2017soft} and an efficient approximate computation exists~\cite{zhao2024bsq}.

The quantization-based auto-encoder enables compressing complex visual content and generating photorealistic images.
However, the learned visual tokens yield inferior performance on understanding tasks~\cite{team2024chameleon,xie2024showo} because of lacking semantic training objectives.

\myparagraph{Language-Image Pre-training} learns visual representation from natural language supervision via a contrastive objective~\cite{radford2021clip,gutmann2010nce,oord2018infonce}.
The training data is image-text pair $(\mX, \mY)$, where $\mY$ is free-form alt-text or short captions encoded in enumerable text tokens.
We employ a visual encoder $\cE_\mathrm{v}$ and a text encoder $\cE_\mathrm{t}$ to obtain the visual and text embeddings
$\vv=\frac{\cE_\mathrm{v}(\mX)}{\| \cE_\mathrm{v}(\mX) \|_2}$, and
$\vw=\frac{\cE_\mathrm{t}(\mY)}{\| \cE_\mathrm{t}(\mY) \|_2}$.

Given a batch of samples $\sB$, the contrastive loss, such as InfoNCE~\cite{oord2018infonce},  learns to associate embedding pairs for the same sample and separate pairs that are not. 
{\small
\begin{align}
    \cL_\mathrm{align}(\vv, \vw) =\sum_{i=1}^{|\sB|} \left( \log\frac{e^{t\vv_i^\top\vw_i}}{\sum_{j=1}^{|\sB|} e^{t\vv_i^\top\vw_j}} + \log\frac{e^{t\vv_i^\top\vw_i}}{\sum_{j=1}^{|\sB|} e^{t\vv_j^\top\vw_i}} \right).
\label{eq:clip}
\end{align}
}

The contrastive-based alignment leads to strong visual representations, which can be integrated into state-of-the-art LLMs through cheap and fast adaptation for visual-language understanding~\cite{liu2023llava,liu2024llava1.5}.
However, it cannot generate visual content due to the encoder-only design.

\section{Quantized Language-Image Pre-training}
\label{sec:method}

Our goal is a text-aligned visual tokenizer whose visual embeddings are projected in a shared space with the text embeddings.
We start from BSQ-autoencoder and add a contrastive language-image alignment branch.
See Figure~\ref{fig:overview} for an illustration.
Specifically, we use a text encoder $\cE_\mathrm{t}$ to obtain the language feature $\vw$ of alt-text $\mY$ accompanying the input image $\mX$.
In the visual encoder $\cE_\mathrm{v}$, we append a learnable classification token $\vx_\mathrm{cls}$ and obtain an extra latent embedding $\vz_\mathrm{cls}$ through $\cE_\mathrm{v}$.
$(\mZ, \vz_\mathrm{cls}) =\cE(\mX; \vx_\mathrm{cls})\in\R^{\left(\frac{H}{p}\times\frac{W}{p}+1\right)\times d}$.
The normalized global visual feature for alignment is computed through a linear projection head $h_\mathrm{v}$:
$\vv=\frac{h_\mathrm{v}(\vz_\mathrm{cls})}{\| h_\mathrm{v}(\vz_\mathrm{cls}) \|_2}$.
Though it seems straightforward at first glance, we observe several challenges when training \ours and elaborate on how we handle them as follows.

\begingroup
\setlength{\intextsep}{2pt}%
\setlength{\columnsep}{8pt}%
\begin{wrapfigure}{r}{0.6\linewidth}
    \centering
    \includegraphics[width=\linewidth]{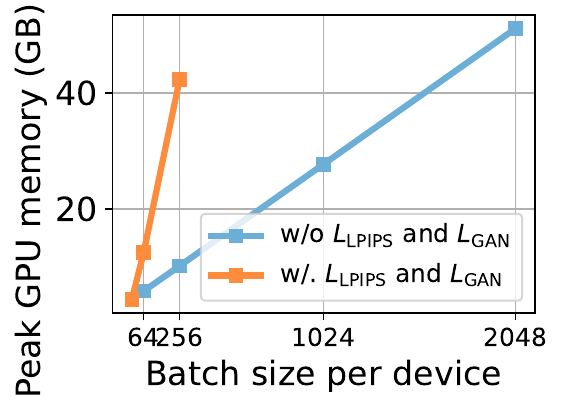}
    \vspace{-25pt}
  \caption{\textbf{Memory usage of \ours.}}
  \label{fig:memory}
\end{wrapfigure}

\myparagraph{Two-stage training.}
Training \ours at once is infeasible.
It is common practice to use a perceptual and adversarial loss for high-quality reconstruction.
Both losses rely on an extra convolutional network~\cite{simonyan2015vgg,karras2020stylegan2} and thus increase the memory footprint (See Figure~\ref{fig:memory}).
On the other hand, effective contrastive learning requires a large batch size (32k$\sim$98k~\cite{zhai2023siglip}).
To reduce memory costs, we opt for a decoupled training recipe in two stages.

\endgroup

\begin{figure}[!bt]
    \centering
    \includegraphics[width=.95\linewidth]{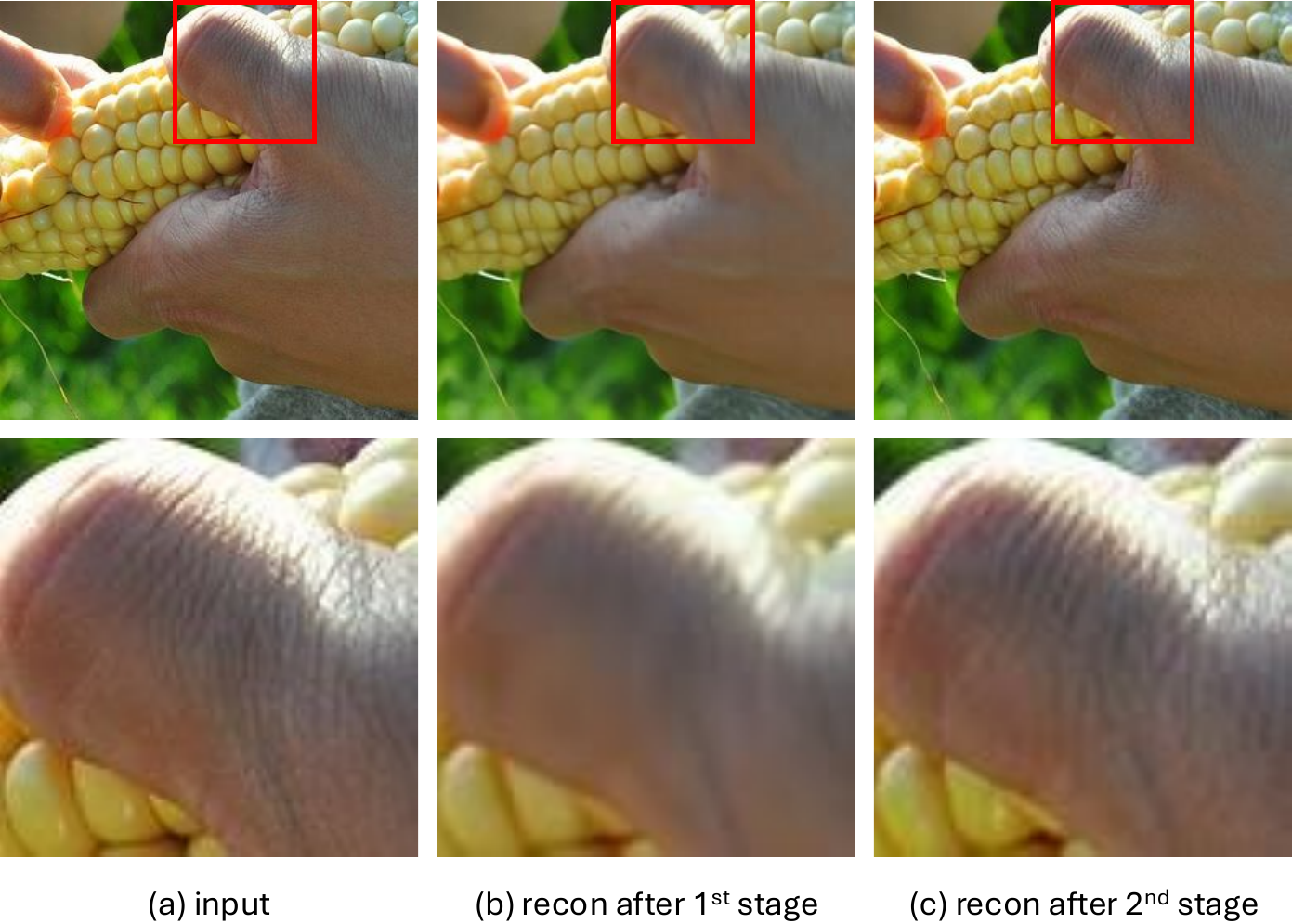}
    \vspace{-5pt}
    \caption{
    \textbf{Comparison of reconstruction results to the input image after the first and second stage.}
    The second-stage model produces more high-frequency details.
    The figure is best viewed on a PDF viewer with zoom-in.
    }
    \label{fig:recon}
\end{figure}

In the \textbf{first} stage, we optimize a weighted sum of reconstruction loss, quantization loss in \Eqsref{eq:bsq}, and contrastive loss in \Eqsref{eq:clip} \emph{without} the perceptual and adversarial loss:
\begin{align}
    \E_{\mX,\mY} \left[ \alpha_{r} \cL_\mathrm{mse} + \alpha_{q}\cL_\mathrm{BSQ} + \alpha_{a} \cL_\mathrm{align}(\vv, \vw) \right].
\end{align}
Here, we prioritize learning semantics-rich representation over better visual reconstruction, which is not always beneficial for representation learning.
We elaborate on our choice of balancing losses in the following paragraph.

In the \textbf{second} stage, we improve the reconstruction quality and restore higher-frequency details by fine-tuning the quantization bottleneck and the visual decoder:
\begin{align}
    \E_{\mX}\left[\alpha_r' \cL_\mathrm{mse}
    + \alpha_q' \cL_\mathrm{BSQ}
    + \alpha_p' \cL_\mathrm{LPIPS}
    + \alpha_g' \cL_\mathrm{GAN}
    \right],
\end{align}
where $\alpha_r'=\alpha_q'=1$, and $\alpha_p'=\alpha_g'=0.1$.
We drop the text encoder and freeze the visual encoder to prevent degradation when the batch-size restriction is relaxed.
See Figure~\ref{fig:recon} for the reconstruction result after two stages.

\myparagraph{Accelerated training with better initializations.}
Training a visual tokenizer with only a reconstruction objective is data efficient\footnote{A common recipe is to train on ImageNet-1K for 100 epochs. In other words, the model sees 1.3 billion samples}. 
In contrast, CLIP-style training requires 30$\sim$50 billion samples to maximize performance.
To narrow the gap, we propose to initialize the visual encoder from either Masked Image Modeling (MIM) pre-training~\cite{fang2024eva02} or contrastive language-image pre-training (CLIP) and the text encoder from CLIP.
Empirically, this significantly increases convergence and training can be finished using 4 billion samples, 10$\times$ faster than training from scratch. 

\begingroup
\setlength{\intextsep}{2pt}%
\setlength{\columnsep}{8pt}%
\begin{wrapfigure}{r}{0.58\linewidth}
    \centering
    \includegraphics[width=\linewidth]{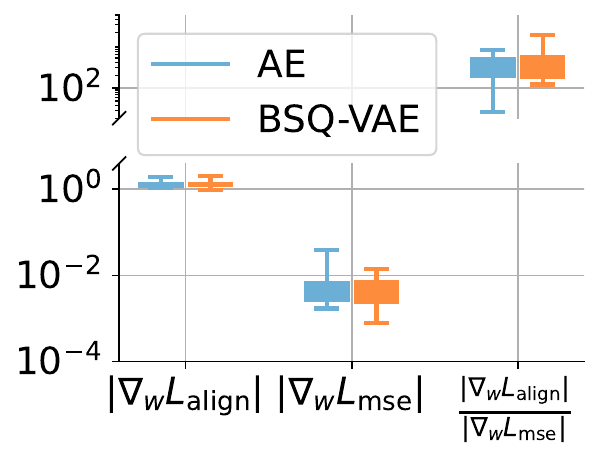}
    \vspace{-25pt}
  \caption{
  \textbf{Comparison of gradient magnitude.}
  Here, $\vw$ refers to the linear layer in the visual encoder's last MLP.
  }
  \label{fig:grad}
\end{wrapfigure}

\myparagraph{Balancing reconstruction and alignment objectives}.
It is important to balance the reconstruction and alignment objective, namely $\alpha_r:\alpha_a$.
If we probe the gradient of each loss with respect to the last shared layer,~\ie the linear layer in the visual encoder's last MLP, we see a difference of several orders of magnitude, leading to different convergence rates between the alignment and reconstruction objectives.
The problem seems more distinct when the straight-through estimator~\cite{bengio2013ste} exists.
We visualize this phenomenon in Figure~\ref{fig:grad} by comparing the gradient norm of two AEs, one of whose quantization bottleneck is replaced with an identity mapping without compression.
To mitigate this, we propose a \emph{post-hoc} way to weigh the two terms.
Specifically, we first train the model with either reconstruction or alignment loss only and then choose the multi-task loss weight to be inversely proportional to
the final loss values,~\ie
${\alpha_r}/{\alpha_a}\approx{\cL_\mathrm{align}(\infty)}/{\cL_\mathrm{mse}(\infty)}$,
where $\cL_{(\cdot)}(\infty)$ denotes the loss value after convergence.

\endgroup

We opt out of adaptive weight methods~\cite{chen2018gradnorm,esser2021vqgan,sener2018moo} for the reason below.
Adaptive weight tuning requires computing the gradient with respect to the last shared layer in the visual encoder.
Therefore, we need an additional backward call of the decoder which introduces non-negligible ($\sim$$\frac{1}{3}$) time and memory overhead.
In our experiments, we find the ratio determined above is robust and works well for different settings of patch size and model parameters.

\myparagraph{Improved bottleneck in BSQ-AE.}
In addition to the training recipe, we improve the tokenizer by replacing
linear projection from the latent space $\vz\in\R^{d}$ to the codebook space $\vu\in S^{L-1}$ with an MLP.
So is the mapping from $\hat{\vu}$ to $\hat{\vz}$ symmetrically.
\begin{align}
\vu =\mathrm{MLP}_\Downarrow(\vz),
\hat{\vu} = \frac{1}{\sqrt{L}}\sign(\vu),
\hat{\vz} = \mathrm{MLP}_\Uparrow(\hat{\vu}),
\end{align}
where $\mathrm{MLP}_{\Downarrow/\Uparrow}$ denotes down/up projection respectively.
Since now the quantization bottleneck is deeper, we optionally add an auxiliary term $\| \sg(\hat{\mZ}) - \mZ \|_2$ during training similar to the commitment loss in VQ-VAE~\cite{van2017vqvae}.
Though it was not necessary in the linear case~\cite{zhao2024bsq}, we see adding it improves reconstruction in our case.

\subsection{Unifying Understanding and Generation}
\label{sec:method:um3}

Now that we have visual tokens aligned with language, we concatenate them with text tokens with appropriately padded special tokens.
On top of this visual-textual token sequence, we apply a Transformer to predict the next token in an auto-autoregressive way without bells and whistles to see if it generates multiple modalities.
See Figure~\ref{fig:overview} (c).
We call our final model the Unified Multimodal Model (\ourlm).

\myparagraph{Architecture.}
We begin with the Llama 3 architecture~\cite{dubey2024llama3}.
To handle the issue of norm growth due to competition from multiple modalities reported by Chameleon~\cite{team2024chameleon}, we apply query-key normalization (QK-Norm)~\cite{dehghani2023vit22b} in the attention layer.
We observe adding QK-Norm is compatible with a pre-trained Llama 3 without QK-Norm.
Therefore, instead of training from scratch like Chameleon, we start from Llama 3 initialization which greatly accelerates training.
We augment the token embedding and the output layers to fit the visual tokens.
The augmented part is initialized with the mean of the existing text embeddings $  \ve_i = \left(\sum_{j=1}^{V_\mathrm{t}} \ve_j \right)/{V_\mathrm{t}}, \forall i \in [V_\mathrm{t}+1, V_\mathrm{t}+V_\mathrm{v}]$, where $V_\mathrm{t}$ and $V_\mathrm{v}$ denotes the vocabulary size of textual and visual tokens.

To alleviate the logit shift problem, we apply the softmax to textual and visual tokens separately:
\begin{align}
    \sum_{i=1}^{V_\mathrm{t}+V_\mathrm{v}} \left( 1_{[i\leq V_\mathrm{t}]} \log\frac{e^{x_i}}{\sum_{j=1}^{V_\mathrm{t}}e^{x_j}} + 1_{[i> V_\mathrm{t}]} \log\frac{e^{x_i}}{\sum_{j=V_\mathrm{t}+1}^{V_\mathrm{t}+V_\mathrm{v}}e^{x_j}} \right).
\end{align}

\myparagraph{Data Mixing.}
Each mini-batch is a mixture of text-only, image-text, or text-image.
Inspired by the warm-up schedule for learning rate~\cite{goyal2017accurate}, we propose a \emph{calm-down} schedule for mixing data,~\ie the proportion of text-only data in a mini-batch linearly decays from $r_0$ to $r_T$ with respect to training step $t$:
\begin{align}
    r(t) = 
    \begin{cases}
    \frac{r_T-r_0}{T}(t-T) + r_T, & \text{if } t\leq T \\
    r_T,              & \text{otherwise}
    \end{cases},
\end{align}
where $r_0,r_T$ are pre-defined hyper-parameters and $0<r_T<r_0$.
This prevents the language modeling ability from collapsing at the beginning of multi-modality training.

\section{Experiments}
\label{sec:exp}

\subsection{Datasets}

\begin{table}[!bt]
    \centering
    \tablestyle{2pt}{1.05}
    \resizebox{\linewidth}{!}{
    \begin{tabular}{cccc}
    \toprule
    Dataset  & Images & Text (\# tok/src) & Usage/Metrics \\
    \midrule
    DataComp-1B~\cite{gadre2023datacomp} & 1B & 20B/alt-text & \ours \\
    LAION-COCO~\cite{laioncoco}\tablefootnote{\scriptsize{\url{hf.co/datasets/guangyil/laion-coco-aesthetic}}} & 4M\scriptsize{\textcolor{gray}{/600M}} & 40M/BLIP2 & T2I (LlamaGen), \ourlm \\
    SA-1B~\cite{kirillov2023sam} & 11M & 400M/\scriptsize{Qwen2VL-7B} & T2I (LlamaGen), \ourlm \\
    CC-12M~\cite{changpinyo2021cc12m} & 6M\scriptsize{\textcolor{gray}{/12M}} & 200M/\scriptsize{Qwen2VL-7B} & \ourlm \\
    DCLM~\cite{li2024dclm} & - & 300B/raw+filtered & \ourlm \\
    LAION-CC-SBU~\cite{liu2024llava1.5} & 558K & -/BLIP2 & VLM (LLaVA-1.5) \\
    LLaVA-Instruct~\cite{liu2024llava1.5} & 665K & -/convo. & VLM (LLaVA-1.5) \\
    ImageNet~\cite{deng2009imagenet} & 1.3M & -/label & Classi. (ZS), Recon. (RC) \\
    MS-COCO~\cite{lin2014mscoco,chen2015cococaption} & 160K & 10M/MTurk & Caption, generation \\
    \bottomrule
    \end{tabular}
    }
    \vspace{-5pt}
    \caption{\textbf{Dataset summary.}
    We list the statistics of datasets used throughout the paper, including the number of images, the number of text tokens with source, and the usage of the respective dataset.
    }
    \label{tab:dataset}
\end{table}

Table~\ref{tab:dataset} summarizes our datasets.
To train \ours, we use \textbf{DataComp-1B}~\cite{gadre2023datacomp}, the largest public image-text pair dataset with 1B samples.
Training details are in Sec.~\ref{sec:supp:details}.
We evaluate the understanding and reconstruction performance on the validation set of ImageNet-1k~\cite{deng2009imagenet}.

For vision-language understanding, we use the pre-training and instruct-tuning data from LLaVA 1.5~\cite{liu2024llava1.5}.
The evaluation benchmarks will be covered in Sec~\ref{sec:exp:evaluate}. 

For text-to-image generation, we use images from Conceptual 12M (CC-12M)~\cite{changpinyo2021cc12m},
SA-1B~\cite{kirillov2023sam}, and a 5M subset of LAION-COCO~\cite{laioncoco} filtered by aesthetic scores.
We use Qwen2-VL-7B~\cite{wang2024qwen2vl} to generate captions and use FLAN-T5~\cite{raffel2020t5,chung2024flant5} to obtain the text embeddings for conditioning.

To train the unified multi-modal model for understanding and generation, we use a mixture of text data from DCLM-baseline~\cite{li2024dclm} (a 300B-tokens subset), image-text pairs from CC-12M+SA-1B (18M images, or 10B tokens in total).

\subsection{Evaluating \ours}
\label{sec:exp:evaluate}

We validate the effectiveness of \ours on a wide spectrum of visual and multi-modal benchmarks.
We categorize them into three parts,~\ie vision-centric understanding, vision-language understanding, and text-conditioned visual generation.
Finally, we showcase the performance of \ourlm on a combination of text-only, I2T, and T2I tasks.

\myparagraph{Vision-centric understanding} includes (1) image classification, measured by zero-shot accuracy and linear-probing accuracy, and (2) reconstruction quality, measured by reconstruction FID (rFID)~\cite{heusel2017fid}, PSNR, and SSIM~\cite{wang2004ssim}.

\myparagraph{Vision-language understanding}
takes as input one or more images $\mX$ and a text sequence $\mY_i$, often known as a prompt or an instruction, and outputs another text sequence $\mY_o$ that follows the prompt.
Following LLaVA 1.5~\cite{liu2024llava1.5}, we employ \ours's visual encoder $\cE_\mathrm{v}$ on the image, adapt the visual embeddings through a learnable projection network $\cF_\mathrm{proj}$, and feed the adapted feature into a pre-trained LLM.
{\small
\begin{align}
    \mH_\mathrm{v} = \cF_\mathrm{proj}(\cE_\mathrm{v}(\mX)), \quad \mY_o \sim \mathrm{LLM}(\mH_\mathrm{v};\mY_i).
\end{align}
}
Instruction tuning undergoes two stages: (1) feature alignment, where we train the visual-to-text projector, and (2) end-to-end fine-tuning, where we train the projector and LLM using curated instruction-following data.
We evaluate the instruction-tuned model on visual question-answering datasets including VQAv2~\cite{goyal2017vqav2}, GQA~\cite{hudson2019gqa}, TextVQA~\cite{singh2019textvqa}, plus more comprehensive VLM benchmarks including POPE~\cite{li2023pope}, MME~\cite{fu2023mme}, and MM-Vet~\cite{yu2023mmvet}.

\myparagraph{Text-conditioned Image Generation} (T2I) takes as input a short caption $\mY_i$ and outputs an image $\mX$ that depicts the text description.
We employ \ours to transform the input image into a set of discrete visual token indices $\{ k_1, \cdots, k_N\}$, where $N=HW/p^2$, and use a text encoder to convert the caption into text embeddings $\cE_\mathrm{t}(\mY_i)$.
A Llama-2 style Transformer~\cite{touvron2023llama2} learns from scratch the visual token sequence auto-regressively with the adapted textual embedding as the prefix condition. 
{\small
\begin{align}
    \mH_\mathrm{t}=\cG_\mathrm{proj}(\cE_\mathrm{t}(\mY_i)), \quad k_n \sim p(k_n' | \mH_\mathrm{t}, k_{<n}).
\end{align}
}

\myparagraph{Unified Multimodal Models.}
We evaluate \ourlm on a suite of language-only benchmarks, image-to-text captioning, and text-to-image generation.
The language-only benchmarks include ARC-Challenge~\cite{clark2018arcc}, HellaSwag~\cite{zellers2019hellaswag}, PIQA~\cite{bisk2020piqa}, Social IQA~\cite{sap2019socialiqa}, and WinoGrande~\cite{sakaguchi2021winogrande}.
For captioning, we report BLEU@4, METEOR, and CIDEr on the MS-COCO Karpathy split.
For T2I generation, we report generation FID      and CLIPScore~\cite{hessel2021clipscore} on MS-COCO 30k.

\begin{table}[!tb]
    \centering
    \tablestyle{2pt}{1.05}
    \resizebox{\linewidth}{!}{
    \begin{tabular}{lcHcccccc}
    \toprule
    \multicolumn{2}{c}{} & \# Param. &  0-shot & & Comp. & \multicolumn{3}{c}{Reconstruction} \\ 
    & Seen Data  & ($|\cE|+|\cG|+|\cQ|$) &  Acc.\higherbetter & \# bits & Ratio & rFID\lowerbetter & PSNR\higherbetter & SSIM\higherbetter \\
    \midrule
    \multicolumn{8}{l}{\textsc{(Base backbone)}} \\
    CLIP~\cite{radford2021clip} & WIT-400M &  87M+0+0 & 68.3 & / & / & / & / & / \\
    EVA-CLIP~\cite{sun2023evaclip} & Merged-2B & 87M+0+0 & 74.7 & / & /  & / & / & / \\
    SigLIP-B~\cite{zhai2023siglip} & WL-10B & 87M+0+0 & 76.7 & / & /  & / & / & / \\
    VQGAN~\cite{esser2021vqgan} & IN-1k  & 29M+42M+4M &  / & 14 & 438.8  & 4.98 & - & - \\
    MaskGIT~\cite{chang2022maskgit} & IN-1k & 24M+30M+6k & / & 10 & 614.4  & 1.98 & 18.63 & 0.4619 \\
    MoVQGAN~\cite{zheng2022movq} & IN-1k  & (82.7M) & / & $^\&$40~~~ & 153.6 & 1.12 & 22.42 & 0.6731 \\
    RQ-VAE/f32~\cite{lee2022rqvae} & IN-1k & & / & $^\&$112~~~~~ & 219.4 & 2.69 & - & - \\
    \compare{}OpenCLIP-B~\cite{cherti2023openclip} & \compare{}DC-1B  & \compare{}87M+0+0 & \compare{}73.5 & \compare{}/ & \compare{}- & \compare{}/ & \compare{}/ & \compare{}/ \\
    \compare{}BSQViT~\cite{zhao2024bsq}$^\dagger$ & \compare{}DC-1B &  87M+87M+1M & \compare{}/ & \compare{}28 & \compare{}219.4 & \compare{}3.81 & \compare{}24.12 & \compare{}0.6638 \\
    \compare{}\ours-B (ours) & \compare{}DC-1B & \compare{}87M+87M+1M  & \compare{}74.3  & \compare{}28 & \compare{}219.4 & \compare{}3.21 &  \compare{}23.16 & \compare{}0.6286 \\
    \midrule
    \multicolumn{8}{l}{\textsc{(Base backbone, Smaller patch)}} \\
    SigLIP-B~\cite{zhai2023siglip} & WL-10B & 87M+0+0 & 79.2 & / & / & / & / & / \\
    DALL-E dVAE~\cite{ramesh2021dalle} & \scriptsize{CC3M+YF} & 54M+44M+0 & / & 13 & 118.2 & 32.63 & 27.31 & 0.7943 \\ 
    ViT-VQGAN~\cite{yu2022vitvqgan} & IN-1k & 91M+91M+0.5M & / & 13 & 118.2 & 1.55 & - & - \\
    SD-VAE 1.x~\cite{rombach2022ldm} & OI-2M &  & / & 14  & 109.7 & 1.40 & 23.65 & 0.6354 \\
    SD-VAE 2.x~\cite{podell2023sdxl} & \scriptsize{OI-2M+LAae} & & / & $^\#$64~~~ & 24~~ & 0.70 & 26.90 & 0.7592 \\
    SDXL-VAE~\cite{podell2023sdxl} & \scriptsize{OI-2M+LAae++} & & / & $^\#$64~~~ & 24~~ & 0.67 & 27.37 & 0.7814 \\
    SBER-MoVQGAN~\cite{sber2023movqgan} & \scriptsize{LAHR-166M} & 29M+42M+4M & / & 14 & 109.7 & 0.96 & 26.45 & 0.7250 \\ 
    BSQViT~\cite{zhao2024bsq} & IN-1k & 87M+87M+28k & / & 18 & ~~85.3 & 0.99 & 27.78 & 0.8171 \\
    \compare{}EVA-CLIP~\cite{sun2023evaclip}$^\dagger$ & \compare{}DC-1B & \compare{}87M+0+0 & \compare{}77.2 & \compare{}/ & \compare{}/ & \compare{}/ &\compare{} / & \compare{}/ \\
    \compare{}\ours-B (ours) & \compare{}DC-1B & \compare{}87M+87M+1M & \compare{}75.6 & \compare{}28 & \compare{}~~54.8 & \compare{}0.70 & \compare{}26.79 & \compare{}0.7905 \\
    \midrule
    \multicolumn{8}{l}{\textsc{(Large backbone)}} \\
    CLIP/f14~\cite{radford2021clip} & WIT-400M & 304M+0+0 & 75.5 & / & / & / & / & / \\
    SigLIP-L~\cite{zhai2023siglip} & WL-10B & 304M+0+0 & 80.5 & / & / & / & / & / \\
    OpenCLIP-L~\cite{cherti2023openclip} & DC-1B & 304M+0+0 & 79.2 & / & / & / & / & / \\
    EVA-CLIP-L~\cite{sun2023evaclip} & Merged-2B & 304M+0+0 & 79.8 & / & / & / & / & / \\
    Open-MAGVIT2~\cite{yu2024magvit2,luo2024openmagvit2} & IN-1k & 50M+65M+18k  & / & 18 & ~~85.3 & 1.17 & 21.90 & - \\
    VILA-U~\cite{wu2024vilau} & \scriptsize{WL-10B+CY-1B} & 316M+42M+134M & 73.3 & $^\&$56~~~ & ~~27.4 & 1.80 & - &  - \\
    \midrule
    \multicolumn{8}{l}{\textsc{(Large backbone, high resolution)}} \\
    CLIP/f14~\cite{radford2021clip} & WIT-400M &  304M+0+0 & 76.6 & / & / & / & / & / \\
    SigLIP-L~\cite{zhai2023siglip} & WL-10B & 304M+0+0 & 82.1 & / & / & / & / & / \\
    \compare{}EVA-CLIP-L~\cite{sun2023evaclip} & \compare{}Merged-2B & \compare{}304M+0+0 & \compare{}80.4 & \compare{}/ & \compare{}/ & \compare{}/ & \compare{}/ & \compare{}/ \\    
    \compare{}VILA-U~\cite{wu2024vilau} (SO400M) & \compare{}\scriptsize{WL-10B+CY-1B} & \compare{}428M+42M+537M & \compare{}78.0 & \compare{}$^\&$224~~~~ & \compare{}21 & \compare{}1.25 & \compare{}- &  \compare{}- \\
    \compare{}\ours-L (ours) & \compare{}DC-1B &  \compare{}304M+304M+2M & \compare{}79.1 & \compare{}28 & \compare{}168~~ & \compare{}1.46 & \compare{}25.36 & \compare{}0.6903 \\
    \bottomrule
    \end{tabular}
    }
    \vspace{-5pt}
    \caption{\textbf{Comparison to state-of-the-art visual encoders or tokenizers.}
    We \hl{highlight} rows that are most comparable in each group.
    $^\dagger$: our reproduction.
    $^\#$: effective number of bits when latents are stored in \texttt{bf16}.
    $^\&$: quantizer uses residual quantization (RQ), where the total bits are multiplied by RQ depth.
    }
    \label{tab:comparable_tokenizers}
    \vspace{-10pt}
\end{table}

\begin{table*}[!tb]
    \centering
    \resizebox{0.9\linewidth}{!}{
    \subcaptionbox{Balancing Loss.\label{tab:weight}}{
    \begin{tabular}{l|ccc}
    \toprule
    $\alpha_a:\alpha_r$ & ZS\tiny{(\%)} & RC\tiny{(rFID)}\lowerbetter & RC\tiny{(PSNR)} \\
    \midrule
    $1:0$ & 75.7 & 367.8~~ & 11.7 \\
    $1:1$  & 75.1 & 162.6~~ & 17.8 \\
    $1:10^2$  & 74.7 & 41.7 & 22.5 \\
    \default{}$1:10^3$  & \default{}74.3 & \default{}35.3 & \default{}24.5 \\
    $1:10^4$  & 35.4 & 35.6 & 24.5 \\
    $0:1$     & ~~0.1 & 35.7 & 24.5 \\
    \bottomrule
    \end{tabular}
    }
    \subcaptionbox{Initialization.\label{tab:initialization}}{
    \tablestyle{2pt}{1.05}
    \begin{tabular}{l|ccc}
    \toprule
    Pretrain  & ZS\tiny{(\%)} & RC\tiny{(rFID)} & RC\tiny{(PSNR)} \\
    \midrule
    None      &  26.4 & 35.0 & 24.8 \\
    \default{}MIM~\cite{fang2024eva02}  & \default{}74.3 & \default{}35.3 & \default{}24.5 \\
    CLIP~\cite{sun2023evaclip}   &  74.7 & 41.7 & 23.9 \\
    \bottomrule
    \end{tabular}
    }
    \subcaptionbox{Training Recipe.\label{tab:training_recipe}}{
    \tablestyle{2pt}{1.05}
    \begin{tabular}{ll|ccc}
    \toprule
    & & ZS\tiny{(\%)} & RC\tiny{(rFID)} & RC\tiny{(PSNR)} \\
    \midrule
    \default{} & \default{}(1) $\cE_\mathrm{t}$, $\cE_\mathrm{v}$, $\cQ$, $\cG$ & \default{} & \default{}35.3 & \default{}24.49 \\
    \default{}\multirow{-2}{*}{Recipe 1} & \default{}(2) Finetune $\cG$ & \default{} \multirow{-2}{*}{74.3} & \default{}3.21 & \default{}23.16 \\
     & (2)$^*$ (on IN-1k) & & 2.90 & 23.33 \\
    \midrule
    \multirow{3}{*}{Recipe 2} & (1) $\cE_\mathrm{t}$, $\cE_\mathrm{v}$, $\cG$ & \multirow{3}{*}{75.0} & 17.2 & 26.72 \\
    & (2) Train $\cQ$ & & \multirow{2}{*}{13.7} & \multirow{2}{*}{23.34} \\
    & ~~ + Finetune $\cG$ & & & \\
    \bottomrule
    \end{tabular}
    }
    }
    \vspace{-5pt}
    \caption{
    \textbf{Ablation studies of training \ours.}
    ZS: zero-shot classification; RC: reconstruction.
    We \hl{highlight} the default setting.
    }
    \label{tab:ablation}
\end{table*}

\begin{table*}[!tb]
  \vspace{-5pt}
  \centering
  \resizebox{0.9\linewidth}{!}{
  \tablestyle{3pt}{1.05}
  \begin{tabular}{llll|lllllHl}
    \toprule
    Method & Vision Encoder & Res & LLM & VQAv2 & GQA & TextVQA & POPE & MME & SEED & MM-Vet \\
    \midrule
    SEED-X~\cite{ge2024seedx} & ViT-bigG-14 & 448 & LLaMA-2-13B & - & 47.9~~ & - & 84.2 & 1435.7 &  & - \\
    LaVIT~\cite{jin2024lavit} & ViT-G & 224 & LLaMA-2-7B & 68.2~~ & 48.0~~ & - & - & - & - & - \\ 
    EVE~\cite{diao2024eve} & - & 1344 & Vicuna-1.5-7B &   78.6$^{*}$ & 62.6$^{*}$ & 56.8 & 85.0 & 1305.7 & 56.8 & 25.7 \\
    Fuyu & - & 1080 & Persimmon-8B & 74.2~~ & - & - & 74.1 &  ~~728.6 & - & 21.4 \\
    VILA-U~\cite{wu2024vilau} & SigLIP-SO400M & 384 & LLaMA-2-7B & 79.4$^{*}$ & 60.8$^{*}$ & 60.8 & 85.8 & 1401.8 & 59.0 & 33.5 \\
    Chameleon~\cite{team2024chameleon} & VQ-VAE & 512 & LLaMA-2-34B$^{+}$ & 69.6~~ & - & - & - & - & - & - \\
    Show-o~\cite{xie2024showo} & MAGVIT-v2 & 256 & Phi-1.5-1.3B & 59.3$^{*}$ & 48.7$^{*}$ & - & 73.8 & ~~948.4 & - & - \\
    Emu3~\cite{wang2024emu3} & MoVQGAN & 512 & LLaMA-2-8B$^{+}$ & 75.1$^{*}$ & 60.3$^{*}$ & 64.7 & 85.2 & - & 68.2 & 37.2 \\
    \midrule
    \multirow{3}{*}{LLaVA-1.5~\cite{liu2024llava1.5}} & CLIP-Large (orig.) & 336 & \multirow{3}{*}{Vicuna-1.5-7B} & 78.5$^{*}$ & 62.0$^{*}$ & 58.2 & 85.9 & 1510.7 & 58.6 & 30.5 \\
     & CLIP-Large (repro.) & 392 & & 79.1$^{*}$\basenum{(+0.0)} & 62.3$^{*}$\basenum{(+0.0)} & 55.4\basenum{(+0.0)} & 87.5\basenum{(+0.0)} & 1484.9\basenum{(+0.0)} &  & 33.3\basenum{(+0.0)} \\
     & \ours-Large (ours) & 392 &  & 78.3$^{*}$\loss{(-0.8)} & 61.8$^{*}$\loss{(-0.5)} & 55.2\loss{(-0.2)} & 86.1\loss{(-1.4)} & 1498.3\gain{(+13.4)} &  & 33.3\gain{(+0.0)} \\
    \bottomrule
  \end{tabular}
  }
  \vspace{-5pt}
  \caption{
  \textbf{Comparison to vision-language modeling on vision-language understanding benchmarks}.
  \ours's encoder works on par with LLaVA-1.5 with our reproduced CLIP-Large under a controlled experiment.
  }
  \label{tab:vlm_main}
\end{table*}

\subsection{Experiment Results on \ours}
\myparagraph{Main results of tokenizations.}
We compare \ours with the state-of-the-art visual encoders or tokenizers in Table~\ref{tab:comparable_tokenizers}.
\ours-B achieves comparable zero-shot classification accuracy with CLIP-only counterparts.
At the same time, it also enables compression with a similar ratio and decoding with a comparable reconstruction quality.
Specifically, we compare with VILA-U~\cite{wu2024vilau}'s vision tower:
\ours-L with 300M parameters outperforms their shape-optimized ViT (SO) with 400M parameters.
Also, we get a very close rFID while achieving 8$\times$ compression rate than VILA-U. 

Next, we present ablation studies that manifest the advantages of the proposed training strategy.
Note that for efficiency we use ViT-B/16 under a shorter schedule (2 billion seen samples).
Though a shorter schedule may favor single-objective baselines, the conclusions we draw generally hold for a full schedule and a bigger backbone.

\myparagraph{Ablation: how to balance different objectives?}
From Table~\ref{tab:weight}, we see the effect of the loss weights between the alignment and the reconstruction objectives.
At higher $\alpha_a$, the alignment loss takes control and the reconstruction result degrades drastically;
At higher $\alpha_r$, the reconstruction objective dominates and the zero-shot accuracy improves slowly.
With appropriate loss balancing, \ours matches the reconstruction-only model and is close to the CLIP baseline by $\sim$1\% accuracy drop.

\myparagraph{Ablation: How to initialize the visual encoder.}
In Table~\ref{tab:initialization}, we examine different ways of initializing the visual encoder,~\ie (1) random initialization, (2) EVA-02~\cite{fang2024eva02} trained with Masked Image Modeling (MIM) objective on ImageNet-21k, and (3) EVA-CLIP~\cite{sun2023evaclip} trained with CLIP objective on Merged-2B.
We observe poor zero-shot accuracy using random initialization because 2B samples are insufficient for the visual encoder to learn from textual supervision. 
Both MIM and CLIP initializations do not suffer from this and achieve similarly high zero-shot accuracy.
However, MIM works noticeably better at reconstruction than CLIP.
We conjecture that outlier tokens with high norms in CLIP may harm reconstruction~\cite{darcet2024register}.

\myparagraph{Ablation: Two-Stage training.}
In Table~\ref{tab:training_recipe}, we study the two-stage training.
We first show that fine-tuning the visual decoder greatly improves rFID from 35.3 to 3.21 with some loss of PSNR.
Although fine-tuning on ImageNet yields an even better metric, we stick to the original DC-1B images by default because text-to-image generation later needs a more general decoder. 
Next, we explore another stage-wise strategy, where we first train the text-aligned auto-encoder without quantization and only train the quantization while fine-tuning the visual decoder.
We can see an improved zero-shot accuracy and a similar PSNR.
However, the rFID score is much worse than the default recipe where the quantizer is included in the first stage.
Recall that FID measures the distance of the high-level feature extracted from Inception-V3~\cite{szegedy2016inceptionv3}, which are strongly correlated to high-level semantics.
This illustrates the importance of learning quantization with language supervision.

\subsection{Experiment Results on Multimodal Understanding and Generation}

\myparagraph{Main results of VLMs on vision-language understanding.}
We present the performance of VLMs using \ours's encoder on vision-language benchmarks in Table~\ref{tab:vlm_main}.
Since VLM performance varied significantly due to instruction tuning data, model (vision encoder and LLM) size, and the number of visual patches~\cite{laurenccon2024matters},
we tried our best to conduct a \emph{controlled experiment} by strictly following the training data of LLaVA-1.5 and using Vicuna-1.5-7B~\cite{vicuna2023} as the underlying LLM.
As for the vision encoder, we train a CLIP-large with an image resolution of 392 and a patch size of 14 to match \ours.
We see our \ours-equipped VLM works comparably well with our reproduced CLIP-Large baseline.

\myparagraph{Ablation: How to use \ours in VLMs?}
We continue the ablation studies on visual tokenization regarding its effect on vision-language understanding.
Specifically, we replace the vision encoder in LLaVA 1.5 with \ours at different layers. 
We can see that the performance drops severely using the last layer before quantizer $\cQ$ and after $\cQ$, compared to the default second last layer.
For the latter one, we ascribe it to the effect of quantization.
For the first one, the reason could be that the last layer's features focus more on the generative/reconstructing objective due to the skip connection design, leaving features with the highest semantic content to earlier layers~\cite{el2024aim}.
We examine the same auto-encoder model with only the reconstruction objective and see a similar drop, indicating again that the reconstruction-only objective does not provide sufficient semantics.

\begin{table*}[!tb]
    \centering
    \resizebox{0.9\linewidth}{!}{
    \tablestyle{3pt}{1.05}
    \begin{tabular}{lcccccc|ccc|cc}
    \toprule
    & & \multicolumn{5}{c|}{Text-only} & \multicolumn{3}{c|}{I2T (COCO-Karpathy)} & \multicolumn{2}{c}{T2I (COCO-30k)} \\
    Method & \# Params & ARC-C & Hellaswag & PIQA & SIQA & Winogrande &  BLEU@4 & METEOR & CIDEr & gFID\lowerbetter & CLIPScore\higherbetter \\
    \midrule
    Llama-3.2~\cite{dubey2024llama3} & 1.4B & 34.90 & 48.70 & 75.95 & 43.50 & 61.48 & - & - & - & - & - \\
    ZeroCap~\cite{tewel2022zerocap} & 0.5B & - & - & - & - & - & 2.6 & 11.5 & 14.6 & - & - \\
    LlamaGen~\cite{sun2024llamagen} & 0.8B & - & - & - & - & - & - & - & - & 33.4$^{*}$  & 0.257$^{*}$ \\ 
    \ourlm (Ours) & 1.5B & 34.30 & 45.35 & 74.65 & 43.09 & 54.22 & 8.6 & 20.2 & 17.3 & 44.1~~ & 0.250~~ \\
    \bottomrule
    \end{tabular}
    }
    \vspace{-5pt}
    \caption{
    \textbf{Results of the Unified Multi-modal Language Model.}
    The number with $^{*}$ is obtained using the checkpoint trained with a similar number of seen image tokens (60M image samples, or 30B visual tokens) as ours.
    }
    \label{tab:um3}
\end{table*}

\begin{figure*}[!tb]
    \vspace{-5pt}
    \centering
    \includegraphics[width=1.0\linewidth]{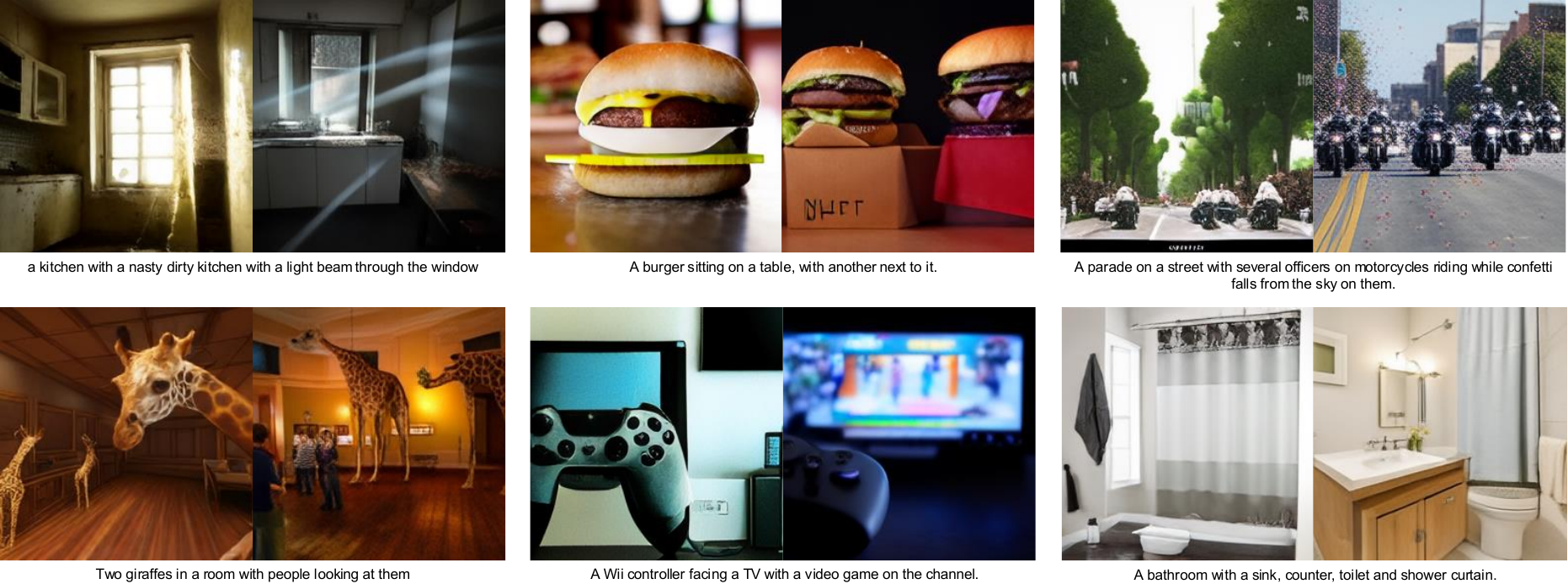}
    \vspace{-15pt}
    \caption{\textbf{Comparison of generated images with conditioning captions in the bottom.}
    For each pair, the left is from LlamaGen+VQGAN and the right is from LlamaGen+\ours-B/16 (ours).
    The caption is also provided at the bottom.
    }
    \label{fig:t2i_vis}
\end{figure*}

\myparagraph{Main results of text-conditioned image (T2I) generation.}
We present the zero-shot image generation result on MS-COCO using 30K captions in Table~\ref{tab:t2i}.
We compare \ours with BSQViT~\cite{zhao2024bsq}, an image tokenizer without semantic alignment, using the same LlamaGen~\cite{sun2024llamagen} framework and show improved generation FID.
Note that \ours is better than the original LlamaGen with VQGAN with only 30\% of the training images.
We also provide results on more comprehensive T2I benchmarks including GenEval~\cite{ghosh2024geneval} and DPG-Bench~\cite{hu2024ella}.
The full comparison is left in Sec.~\ref{sec:supp:geneval}.

\begin{table}[!tb]
    \vspace{-5pt}
    \centering
    \resizebox{\linewidth}{!}{
    \tablestyle{2pt}{1.05}
    \begin{tabular}{lcccc|cccc}
    \toprule
    Contra. & Recon. & \scriptsize{layer\#} & \scriptsize{use $\cQ$} & ZS\tiny{(\%)} & GQA & TextVQA & POPE & MME \\
    \midrule
    \cmark{\tiny{(CLIP-B)}} & \xmark & {-2} &  \xmark & 68.3 & 59.9 & 51.2 & 84.3 & 1397.9 \\
    \midrule
    \cmark & \xmark & {-2} & \xmark & 75.7 & 62.1 & 51.7 & 85.9 & 1411.0 \\
    \midrule
    \multirow{3}{*}{\cmark} & \multirow{3}{*}{\cmark} & \default{}{-2} & \default{}\xmark & \default{}\multirow{3}{*}{74.3} & \default{}61.2 & \default{}51.1 & \default{}86.1 & \default{}1398.7 \\
     &  &  {-1} & \xmark & & 50.7 & 45.0 & 77.2 & 1077.7 \\
     &  &  {-1} & {\cmark} & & 40.4 & 43.0 & 50.6 & 677.17 \\
    \midrule
    {\xmark} & {\cmark}  & {-2} & \xmark & {~~0.1} & 50.8 & 43.8 & 78.3 & 1093.8 \\
    \bottomrule
    \end{tabular}
    }
    \vspace{-5pt}
    \caption{\textbf{Ablations studies on vision-language understanding benchmarks.}
    The first row denotes the original CLIP-B model while all other rows are from our models.
    ``use $\cQ$'' means that the feature is after the quantizer.
    }
    \label{tab:vlm_ablation}
    \vspace{-5pt}
\end{table}

\begin{table}[!tb]
    \vspace{-5pt}
    \centering
    \resizebox{\linewidth}{!}{
    \tablestyle{3pt}{1.05}
    \begin{tabular}{Hlrcccc}
    \toprule
    \multirow{2}{*}{Method} & \multirow{2}{*}{Tokenizer} & \multirow{2}{*}{\# Images} & \multicolumn{2}{c}{MS-COCO 30K} & GenEval & DPG-Bench \\
    & & & gFID\lowerbetter & CLIPScore\higherbetter &  Overall\higherbetter & Overall\higherbetter \\
    \midrule
    \multirow{3}{*}{LlamaGen-0.8B~\cite{sun2024llamagen}} & VQGAN (used in~\cite{sun2024llamagen}) & 50M & 15.68 & 0.309 & 0.32 & 43.22 \\
    & BSQViT-B/16 & 15M & 19.03 & 0.303 & 0.31 & 34.03 \\
    & \ours-B/16   & 15M & \textbf{15.29} & \textbf{0.316} & \textbf{0.48} & \textbf{78.17} \\    
    \bottomrule
    \end{tabular}
    }
    \vspace{-5pt}
    \caption{
    \textbf{Zero-shot generation results on MS-COCO 30K, GenEval~\cite{ghosh2024geneval}, and DPG-Bench~\cite{hu2024ella}.}
    All use LlamaGen-XL~\cite{sun2024llamagen}.
    }
    \label{tab:t2i}
    \vspace{-5pt}
\end{table}

\myparagraph{Qualitative results of T2I generation.}
In Figure~\ref{fig:t2i_vis}, we present side-by-side generated images by LlamaGen with the original VQGAN and \ours.
We put the conditioning caption under each image pair.
We can see images generated by \ours follow captions better by depicting all aspects that might be missing from the VQGAN baseline,~\eg ``light beam'', ``sink, counter'', ``white bush'', and ``people looking at [the giraffes]''.
See Sec.~\ref{sec:supp:generation} for more results.

\myparagraph{Main results of Unified Multimodal Models (\ourlm).}
Finally, we show the performance of the unified multimodal models that perform all text-only, image-to-text, and text-to-image tasks in one \textit{single} model in Table~\ref{tab:um3}.
For reference, we list specialized models with a similar model size. 
For text-only benchmarks, \ourlm achieves comparable results to Llama-3.2 on 3 out of 5 benchmarks.
In zero-shot COCO captioning, \ourlm outperforms ZeroCap~\cite{tewel2022zerocap}, a zero-shot captioning model using CLIP and GPT-2.
In text-conditioned image generation, \ourlm achieves slightly worse gFID but comparable CLIP-Score.

\section{Conclusion}
\label{sec:conclusion}

We present Quantized Language-Image Pre-training, a visual tokenization method that performs well on both understanding and reconstruction.
The visual tokenizer can be seamlessly plugged into state-of-the-art VLMs and image-generation models with comparable performance.
Integrating text-aligned tokens with the pre-trained LLM, we show the feasibility of training a unified multi-modal model.

{
    \small
    \bibliographystyle{ieeenat_fullname}
    \bibliography{main}
}

\clearpage
\setcounter{page}{1}
\appendix

\section{Implementation Details}
\label{sec:supp:details}

\myparagraph{Training \ours.}
Table~\ref{tab:supp:hyperparam:qlip} lists the key hyper-parameters of training \ours-B-8.
The recipe for training other configurations,~\eg \ours-B-16 and \ours-L-14, is similar.

\myparagraph{Training LLaVA.}
This strictly follows the training recipe of LLaVA 1.5 for the sake of a controlled experiment.
For details, please refer to the original paper~\cite{liu2024llava1.5}.

\myparagraph{Training LlamaGen.}
We mostly follow the recipe provided in the original work~\cite{sun2024llamagen}.
Since the authors did not release the training data, we curated the training data by ourselves.
We use a combination of two sources: (1) a 5M subset of LAION-COCO, filtered by aesthetic scores, and
(2) the full set of SA-1B (with 11M images), whose caption is generated by Qwen2-VL-7B~\cite{wang2024qwen2vl}.

\myparagraph{Training \ourlm.}
Table~\ref{tab:supp:hyperparam:um3} lists the key hyper-parameters of training \ourlm-1.5B.

\begin{table}[!tb]
    \centering
    \resizebox{0.9\linewidth}{!}{
    \begin{tabular}{c|c|c}
    config   & Stage 1  & Stage 2 \\
    \midrule
    peak learning rate     &  5e-4  & 5e-4 \\
    $\cE_\mathrm{v}$ learning rate  & 2e-4 &  0 \\
    $\cE_\mathrm{t}$ learning rate  & 2e-5 &  0 \\
    $\cG$ learning rate  & 2e-3 & 1e-4 \\
    learning rate schedule & cosine annealing & cosine annealing \\
    optimizer              & LAMB  & AdamW \\
    optimizer $(\beta_1, \beta_2)$ & (0.9, 0.95) & (0.9, 0.95) \\
    weight decay           & 0.05 & 0.05 \\
    gradient clip          & 5  & 1 \\
    input resolution       & 256 & 256 \\
    patch size             & 8  &  8 \\
    warm-up iterations     & 2,000 & 2,000 \\
    total iterations       & 120,000 & 120,000 \\
    batch size per device  & 512  & 128 \\
    total batch size       & 65,536  & 16,384 \\
    $\cD$ optimizer        & -  & AdamW \\
    $\cD$ learning rate    & -  & 1e-4 \\
    reconstruction loss weight $\alpha_r$ & 1e3 &  1 \\
    contrastive loss weight $\alpha_a$ & 1 &  0 \\
    quantization loss weight $\alpha_q$  & 1 &  1 \\
    perceptual loss weight $\alpha_p$  & 0 &  0.1 \\
    GAN loss weight $\alpha_g$  & 0 &  0.1 \\
    commitment loss weight $\alpha_z$  & 1.0 &  0 \\
    \end{tabular}
    }
    \caption{\textbf{Hyperparamters for training \ours.} Please refer to Sec.\ref{sec:method} for the notions of loss weights.}
    \label{tab:supp:hyperparam:qlip}
\end{table}

\begin{table}[!tb]
    \centering
    \resizebox{0.66\linewidth}{!}{
    \begin{tabular}{c|c}
    config   & Training \ourlm  \\
    \midrule
    peak learning rate     &  1e-4 \\
    learning rate schedule & cosine annealing \\
    optimizer              & AdamW \\
    optimizer $(\beta_1, \beta_2)$ & (0.9, 0.95) \\
    weight decay           & 0.1 \\
    gradient clip          & 1 \\
    warm-up iterations     & 2,000 \\
    total iterations       & 600,000 \\
    batch size per device  & 8  \\
    total batch size       & 512 \\
    sequence length        & 4,096 \\
    calm-down steps        & 10,000 \\
    mix ratio ($r_{\mathrm{text},0}:r_\mathrm{i2t}:r_\mathrm{t2i}$) & 60:1:3 \\
    mix ratio ($r_{\mathrm{text},T}:r_\mathrm{i2t}:r_\mathrm{t2i}$) & 12:1:3 \\
    sampling temperature   & 1.0 \\
    sampling top-$p$       & 0.95 \\
    \end{tabular}
    }
    \caption{\textbf{Hyperparamters for training \ourlm.}}
    \label{tab:supp:hyperparam:um3}
\end{table}

\section{More Results on \ours}
\myparagraph{Full version of Table~\ref{tab:comparable_tokenizers}.}
We present a more detailed comparison to the state-of-the-art visual encoders or tokenizers in Table~\ref{tab:supp:comparable_tokenizers}.
Compared to Table~\ref{tab:comparable_tokenizers}, we add a column that computes the number of parameters.
Though the convolution-based methods,~\eg VQGAN, have fewer parameters than ViT-based methods,~\eg BSQViT and \ours-B, the runtime is slower as is reported in~\cite{zhao2024bsq}.
Therefore, we subsume those under ``base backbone''.

\myparagraph{Linear Evaluation.}
In addition to the zero-shot image classification, we conduct a linear probing evaluation to compare all visual encoder methods.
Table~\ref{tab:supp:hyperparam:linear} gives the linear probing settings.
For VQ-VAE~\cite{van2017vqvae} and LQAE~\cite{liu2023lqae}, we directly copy the numbers from the paper due to the inaccessibility of models.
We see significant improvement in linear classification accuracy over reconstruction-only tokenizers, such as VQ-VAE and BSQ-ViT, and language-quantized tokenizers, such as LQAE.
We explore two probing positions, namely using the reserved \texttt{[CLS]} token (\textit{cls}-token) or the averaged feature tokens (\textit{ft}), and their concatenation.
Using the averaged feature tokens yields a linear probing accuracy similar to the \textit{cls} token, indicating that the encoder learns strong semantics.
As a reference, we also run the linear evaluation on EVA-CLIP~\cite{sun2023evaclip} and see \ours is very close to this upper bound.

\begin{table}[!tb]
    \centering
    \tablestyle{3pt}{1.05}
    \resizebox{0.85\linewidth}{!}{
    \begin{tabular}{lccc}
    \toprule
    Method & Seen Data & Probing Pos. & IN-1k Acc.\tiny{(\%)} \\
    \midrule
    \multicolumn{3}{l}{\textsc{(Base backbone)}} \\
    VQVAE~\cite{van2017vqvae} & IN-1k & / & 18.4 \\
    LQAE~\cite{liu2023lqae} & IN-1k & / & 39.7 \\
    EVA-CLIP-B~\cite{sun2023evaclip} & Merged-2B & \textit{cls}-token & {\textcolor{gray}{82.7}} \\
    BSQViT~\cite{zhao2024bsq}$^\dagger$ & DC-1B & \textit{cls}-token & 29.3 \\
    BSQViT~\cite{zhao2024bsq}$^\dagger$ & DC-1B & \textit{ft} (avg.) & 25.4 \\
    \ours-B (ours) & DC-1B & \textit{cls}-token & 81.8 \\
    \ours-B (ours) & DC-1B & \textit{ft} (avg.) & 77.7 \\
    \ours-B (ours) & DC-1B & \textit{cls} + \textit{ft} & 82.1 \\
    \midrule
    \multicolumn{2}{l}{\textsc{(Large backbone, high resolution)}} \\
    EVA-CLIP-L~\cite{sun2023evaclip} & Merged-2B & \textit{cls}-token  & {\textcolor{gray}{86.3}} \\
    \ours-L (ours) & DC-1B & \textit{cls}-token & 85.2 \\
    \bottomrule
    \end{tabular}
    }
    \caption{\textbf{Linear evaluation on image classification.}
    }
    \label{tab:supp:linear_probe}    
\end{table}

\begin{table}[!tb]
    \centering
    \resizebox{0.8\linewidth}{!}{
    \begin{tabular}{c|c}
    config   & ImageNet linear probing \\
    \midrule
    peak learning rate     & 0.2 / 1.0 (BSQViT) \\
    learning rate schedule & cosine annealing \\
    optimizer              & AdamW \\
    optimizer $(\beta_1, \beta_2)$ & (0.9, 0.999) \\
    weight decay           & 0. \\
    input resolution       & 256 (\ours-B) / 392 (\ours-L) \\
    patch size             & 16 (\ours-B) / 14 (\ours-L)  \\
    warm-up epochs         & 1  \\
    total epochs           & 10 / 20 (BSQViT) \\
    batch size per device  & 128 \\
    total batch size       & 1,024 \\
    \end{tabular}
    }
    \caption{\textbf{Hyperparamters for ImageNet linear probing.}}
    \label{tab:supp:hyperparam:linear}
\end{table}

\begin{figure*}[!tb]
    \centering
    \includegraphics[width=1.0\linewidth]{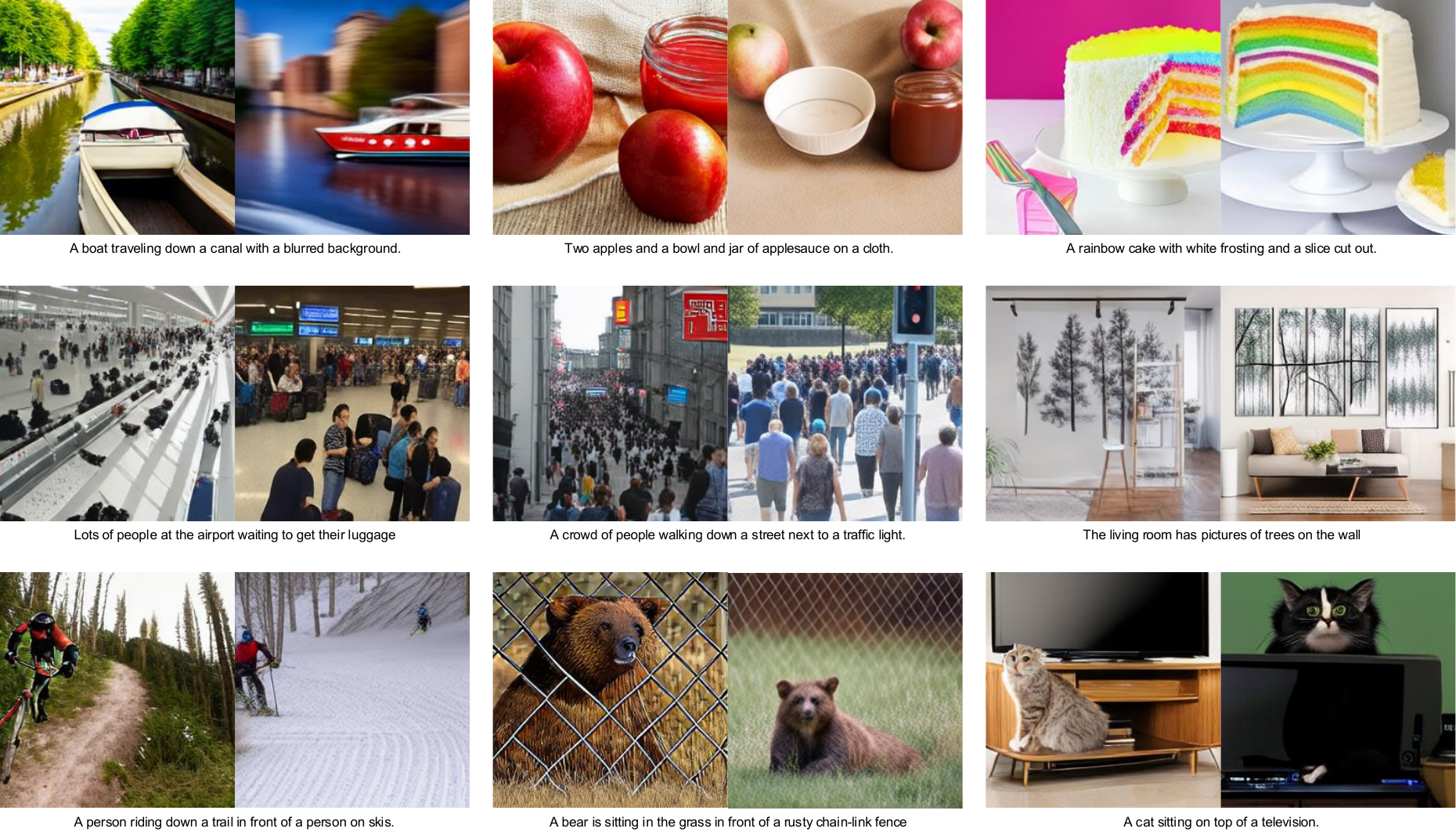}
    \vspace{-15pt}
    \caption{\textbf{Comparison of generated images with conditioning captions in the bottom.}
    For each pair, the left is from LlamaGen+VQGAN and the right is from LlamaGen+\ours-B/16 (ours).
    The caption is also provided at the bottom.
    }
    \label{fig:supp:t2i_vis}
\end{figure*}

\section{More Results on Generation Benchmarks}
\label{sec:supp:geneval}

We show the full results on comprehensive benchmarks such as GenEval~\cite{ghosh2024geneval} and DPG-Bench~\cite{hu2024ella} in Tables~\ref{tab:supp:geneval} and~\ref{tab:supp:DPG-Benchmark} respectively.
Under the same T2I framework, \ours-equipped LlamaGen significantly outperforms the open-sourced VQGAN-LlamaGen and our reproduced baseline with BSQ-ViT.
It also achieves competitive or better results than diffusion-based methods, e.g. SDv1.5 which is trained on much more data. We will add the results in the final version. 

\begin{table}[!thb]
    \centering
    \resizebox{\linewidth}{!}{
    \tablestyle{1pt}{1.05}
    \begin{tabular}{c|l|c|c|c|c|c|c|c}
    \toprule
    Model & Tokenizer &  \scriptsize{Overall} & \scriptsize{Single Obj.} & \scriptsize{Two Obj.} & \scriptsize{Counting} & \scriptsize{Colors} & \scriptsize{Position} & \scriptsize{Attribute} \\
    \midrule
    \multirow{2}{*}{LlamaGen} & VQGAN  & 0.32 & 0.69 & 0.36 & 0.20 & 0.57 & 0.06 & 0.02 \\
    \multirow{2}{*}{(0.8B)} & BSQ-ViT & 0.31 & 0.77 & 0.26 & 0.13 & 0.56 & 0.05 & 0.06 \\
    & \textbf{\ours (Ours)} & \textbf{0.48} & \textbf{0.91} & \textbf{0.59} & \textbf{0.22} & \textbf{0.80} & \textbf{0.17} & \textbf{0.24} \\
    \multicolumn{2}{c|}{{\color{gray} SDv1.5 (0.9B)}} & \color{gray}0.43 & \color{gray}0.97 & \color{gray}0.38 & \color{gray}0.35 & \color{gray}0.76 & \color{gray}0.04 & \color{gray}0.06 \\
    \bottomrule
    \end{tabular}
    }
    \caption{Evaluation on GenEval.}
    \label{tab:supp:geneval}
\end{table}

\begin{table}[!thb]
    \centering
    \resizebox{\linewidth}{!}{
    \tablestyle{2pt}{1.05}
    \begin{tabular}{c|l|c|c|c|c|c|c}
    \toprule
    Model & Tokenizer &  Average & Global & Entity & Attribute & Relation & Other \\
    \midrule
    \multirow{2}{*}{LlamaGen} & VQGAN  & 43.22 & 76.60 & 57.88 & 66.96 & 75.78 & 42.80 \\
    \multirow{2}{*}{(0.8B)} & BSQ-ViT  & 34.03 & 68.39 & 47.70 & 63.40 & 73.77 & 33.60 \\
    & \textbf{\ours (Ours)} & \textbf{78.17} & \textbf{82.37} & \textbf{84.68} & \textbf{86.97} & \textbf{92.50} & \textbf{79.20}  \\
    \multicolumn{2}{c|}{\color{gray}SDv1.5 (0.9B)} & \color{gray}63.18 & \color{gray}74.63 & \color{gray}74.23 & \color{gray}75.39 & \color{gray}73.49 & \color{gray}67.81 \\
    \bottomrule
    \end{tabular}
    }
    \caption{Evaluation on DPG-Bench.}
    \label{tab:supp:DPG-Benchmark}
\end{table}

\section{More Generation Results}
\label{sec:supp:generation}
In Figure~\ref{fig:supp:t2i_vis}, we show more side-by-side generated images by LlamaGen with the original VQGAN and the proposed \ours.
We emphasize the advantage of \ours in terms of better following the captions.
The visual quality can be improved by adding more training data, long training iterations, and larger backbones.
However, this is beyond the scope of this paper.

\begin{table*}[!tb]
    \centering
    \tablestyle{3pt}{1.05}
    \begin{tabular}{lcHcc|cH|ccc}
    \toprule
    \multicolumn{3}{c}{} & \# Param. & & {Understanding} & & \multicolumn{3}{c}{Reconstruction} \\ 
    & Seen Data & Patch Size & ($|\cE|+|\cG|+|\cQ|$) & \# bits & 0-shot Acc.\higherbetter & Lin. Probe\higherbetter & rFID\lowerbetter & PSNR\higherbetter & SSIM\higherbetter \\
    \midrule
    \multicolumn{10}{l}{\textsc{(Base backbone)}} \\
    CLIP~\cite{radford2021clip} & WIT-400M & 16$\times$16 &  87M+0+0 & / & 68.3 & & / & / & / \\
    EVA-CLIP~\cite{sun2023evaclip} & Merged-2B & 16$\times$16 & 87M+0+0 & / & 74.7 & & / & / & / \\
    SigLIP-B~\cite{zhai2023siglip} & WL-10B & 16$\times$16 & 87M+0+0 & / &  76.7 & & / & / & / \\
    VQGAN~\cite{esser2021vqgan} & IN-1k & 16$\times$16 & 29M+42M+4M & 14 & / & & 4.98 & - & -  \\
    MoVQGAN~\cite{zheng2022movq} & IN-1k & 16$\times$16 & (82.7M) & $^\&$40~~~ & / & & 1.12 & 22.42 & 0.6731 \\
    MaskGIT~\cite{chang2022maskgit} & IN-1k & 16$\times$16 & 24M+30M+6k & 10 & / & & 1.98 & 18.63 & 0.4619 \\
    Open-MAGVIT2~\cite{yu2024magvit2,luo2024openmagvit2} & IN-1k & 16$\times$16 & 25M+40M+18k & 18 & / & & 1.53 & 21.53 & - \\
    OpenCLIP-B~\cite{cherti2023openclip} & DC-1B & 16$\times$16 & 87M+0+0 & / & 73.5 &  & / & / & / \\
    BSQViT~\cite{zhao2024bsq}$^\dagger$ & DC-1B & 16$\times$16 & 87M+87M+1M & 28 & / & & 3.81 & 24.12 & 0.6638 \\
    \ours-B (ours) & DC-1B & 16$\times$16 & 87M+87M+1M & 28 & 74.3 & & 3.21 &  23.16 & 0.6286 \\
    \midrule
    \multicolumn{10}{l}{\textsc{(Base backbone, Smaller patch)}} \\
    SigLIP-B~\cite{zhai2023siglip} & WL-10B & 8$\times$8 & 87M+0+0 & / &  79.2 & & / & / & / \\
    DALL-E dVAE~\cite{ramesh2021dalle} & \scriptsize{CC3M+YF} & 8$\times$8 & 54M+44M+0 & 13 & / & & 32.63 & 27.31 & 0.7943 \\ 
    ViT-VQGAN~\cite{yu2022vitvqgan} & IN-1k & 8$\times$8 & 91M+91M+0.5M & 13 & / & & 1.55 & - & - \\
    SD-VAE 1.x~\cite{rombach2022ldm} & OI-2M & 8$\times$8 & 34M+49M+0  & 14 & / &  & 1.40 & 23.65 & 0.6354 \\
    SD-VAE 2.x~\cite{podell2023sdxl} & \scriptsize{OI-2M+LA-ae} & 8$\times$8 & 34M+49M+0 & $^\#$64~~~ & / &  & 0.70 & 26.90 & 0.7592 \\
    SDXL-VAE~\cite{podell2023sdxl} & \scriptsize{OI-2M+LA-ae++} & 8$\times$8 & 34M+49M+0 & $^\#$64~~~ & / &  & 0.67 & 27.37 & 0.7814 \\
    SBER-MoVQGAN~\cite{sber2023movqgan} & \scriptsize{LAHR-166M} & 8$\times$8 & 29M+42M+4M & 14 & / & & 0.96 & 26.45 & 0.7250 \\ 
    BSQViT~\cite{zhao2024bsq} & IN-1k & 8$\times$8 & 87M+87M+28k & 18 &  / & & 0.99 & 27.78 & 0.8171 \\
    EVA-CLIP~\cite{sun2023evaclip}$^\dagger$ & DC-1B & 8$\times$8 & 87M+0+0 & / & 77.2 &  & / & / & / \\
    \ours-B (ours) & DC-1B & 8$\times$8 & 87M+87M+1M & 28 & 75.6 & & 0.70 & 26.79 & 0.7905 \\
    \midrule
    \multicolumn{10}{l}{\textsc{(Large backbone)}} \\
    CLIP/f14~\cite{radford2021clip} & WIT-400M & 14$\times$14 &  304M+0+0 & / & 75.5 & & / & / & / \\
    SigLIP-L~\cite{zhai2023siglip} & WL-10B & 16$\times$16 & 304M+0+0 & / &  80.5 & & / & / & / \\
    OpenCLIP-L~\cite{cherti2023openclip} & DC-1B & 16$\times$16  & 304M+0+0 & / & 79.2 & & / & / & / \\
    EVA-CLIP-L~\cite{sun2023evaclip} & Merged-2B & 16$\times$16  & 304M+0+0 & / & 79.8 &  & / & / & / \\
    Open-MAGVIT2~\cite{yu2024magvit2,luo2024openmagvit2} & IN-1k & 16$\times$16 & 50M+65M+18k & 18 & / & & 1.17 & 21.90 & - \\
    VILA-U~\cite{wu2024vilau} & \scriptsize{WL-10B+CY-1B} & 16$\times$16 & 316M+42M+134M & $^\&$56~~~ & 73.3 & & 1.80 & - &  - \\
    \midrule
    \multicolumn{10}{l}{\textsc{(Large backbone, high resolution)}} \\
    CLIP/f14~\cite{radford2021clip} & WIT-400M & 14$\times$14 &  304M+0+0 & / & 76.6 & & / & / & / \\
    SigLIP-L~\cite{zhai2023siglip} & WL-10B & 16$\times$16 & 304M+0+0 & / &  82.1 & & / & / & / \\
    EVA-CLIP-L~\cite{sun2023evaclip} & Merged-2B & 16$\times$16  & 304M+0+0 & / & 80.4 &  & / & / & / \\
    VILA-U~\cite{wu2024vilau} & \scriptsize{WL-10B+CY-1B} & 14$\times$14 & 428M+42M+537M & $^\&$224~~~~ & 78.0 & & 1.25 & - &  - \\
    \ours-L (ours) & DC-1B & 14$\times$14 & 304M+304M+2M & 28 & 79.1 & & {1.46} & {25.36} & {0.6903} \\
    \bottomrule
    \end{tabular}
    \caption{\textbf{Comparison to state-of-the-art visual encoders/tokenizers.}
    $^\dagger$:our reproduction.
    $^\#$: effective number of bits when latents are stored in \texttt{bf16}.
    $^\&$: quantizer uses residual quantization (RQ), where the total bits are multiplied by RQ depth.
    }
    \label{tab:supp:comparable_tokenizers}    
\end{table*}

\end{document}